\documentclass{article}
\usepackage[utf8]{inputenc}
\usepackage{dirtytalk}
\usepackage{graphicx}
\DeclareGraphicsExtensions{.png,.pdf}

\title{Societal biases reinforcement through machine learning – A credit scoring perspective}
\author{Bertrand K. Hassani}

\begin{document}

\maketitle

\begin{abstract}
Does machine learning and AI ensure that social biases thrive ? This paper aims to analyse this issue. Indeed, as algorithms are informed by data, if these are corrupted, from a social bias perspective, good machine learning algorithms would learn from the data provided and reverberate the patterns learnt on the predictions related to either the classification or the regression intended. In other words, the way society behaves whether positively or negatively, would necessarily be reflected by the models. In this paper, we analyse how social biases are transmitted from the data into banks loan approvals by predicting either the gender or the ethnicity of the customers using the exact same information provided by customers through their applications\footnote{Bertrand K. Hassani\\

Université Paris 1 Panthéon-Sorbonne, CES106 bd de l'Hôpital, 75013 Paris, France\\

University College London Computer Science, 66–72 Gower Street, London WC1E 6EA, UK\\

bertrand.hassani@gmail.com}.\\


\end{abstract}

\section{Introduction}

\noindent According to the Cambridge dictionary\footnote{https://dictionary.cambridge.org/dictionary/english/bias}, a bias implies the \say{action of supporting or opposing a particular person or thing in an unfair way}. These biases might be unconscious i.e., the person with the bias is not aware of it, or worst, this bias might just be the result of conforming to the norm, as norms are behaviors that are self-enforcing at the group level and are not necessarily positive as it is just something followed by the masses. Social biases, to be precise, occur when we unknowingly or deliberately make a judgment about certain individuals, groups, races, opinion, and so on, due to preconceived notions about the group. These can either be positive or negative beliefs and are often instilled in us based on our own culture and environment. Societal biases, in turn, occur when social biases become the norm.\\

\noindent Social biases have been reported in many papers either released by NGOs, for instance see \cite{hegewisch2018gender} 
,\cite{ons2019}, \cite{CreditCards2019}, or academics see \cite{levinson2012implicit} or \cite{hall2016black} among others\footnote{Numerous sources from both Public Policy and Law departments of Universities, in particular, have been identified.}. As reported in the aforementioned papers, it is clear that both gender gaps and ethnicity gaps exist in remuneration; thus, it would not be surprising that these gaps have impacts on consumption, education or access to loans, though this remains to be proved. This is the objective of this paper using data sets, capturing both gender and ethnicity (among other elements), traditionally used for scoring purposes.\\

\noindent As algorithms learn from data, if these are corrupted, from a social bias perspective, not necessarily from a data quality point of view, then, a good machine learning algorithm would learn from the data provided and reverberate the patterns learnt onto the predictions related either to the classifications or the regressions intended. Therefore, if the data sets are capturing the way society behaves whether it is positive or negative (discrimination towards gender, ethnicity, among others), then this would be reflected by the models; for instance, if someone faces discrimination in their workplace, then this is likely to be reverberated in her remuneration and mechanically in her access to loans; and a \say{good} algorithm will naturally score these discriminated people at a lower level.\\

\noindent Essentially, machine learning captures all features characterizing a phenomenon and then relies on them to make predictions. However, these features may characterize not only the intended phenomenon but might also be informative in characterizing other phenomena, categories of features, or classes. For instance, someone with a low income may be related to the fact that her loan request is not approved, but may also reflect that they are relatively poor, having a \say{blue collar} job as well as their gender, ethnicity, location of their home, and so on. In other words, social biases are naturally and mechanically captured in data, and therefore, we believe that these might be captured and replicated by machine learning algorithms. If these algorithms are used for loan approval or credit scoring purposes, then they might not only replicate these biases but may also validate them as normal and transmit them over time. Indeed, the newest data would be subsequently used to assess future customers' requests. Furthermore, financial institutions profit generating paradigm and regulations have the negative effect of ensuring that the system cannot self correct. Certainly, prudential rules do not allow financial institutions to take more risk than what is prescribed (\cite{witzany2017credit}).\\

\noindent Therefore, in this paper we intend to address the following. Assuming that a credit scoring model is not socially biased, the data used to assess the suitability of a loan applicant should not give any information regarding either their gender or their ethnicity. However, we have seen from numerous reports that the world thrives with inequality, for instance, inequality in remuneration. It is generally accepted that income is one of the main elements for a bank to accept a loan request. Therefore, in this paper, using credit data, we will try to predict both the ethnicity of the applicants and their gender. We assume that if we are able to predict either of these from a data set used for credit scoring, then it means that the intrinsic characteristics of each population will be spilled over on their access to loans. Thus, it would be necessary to correct the rating for the bias identified while ensuring that regulatory requirements are fulfilled.\\

\noindent In this paper, after presenting the data sets, we will introduce the methodology and discuss the results obtained. A last section offers a conclusion.

\section{The Data Sets}

\begin{figure}[h!]
\centering
\includegraphics[width=1\textwidth]{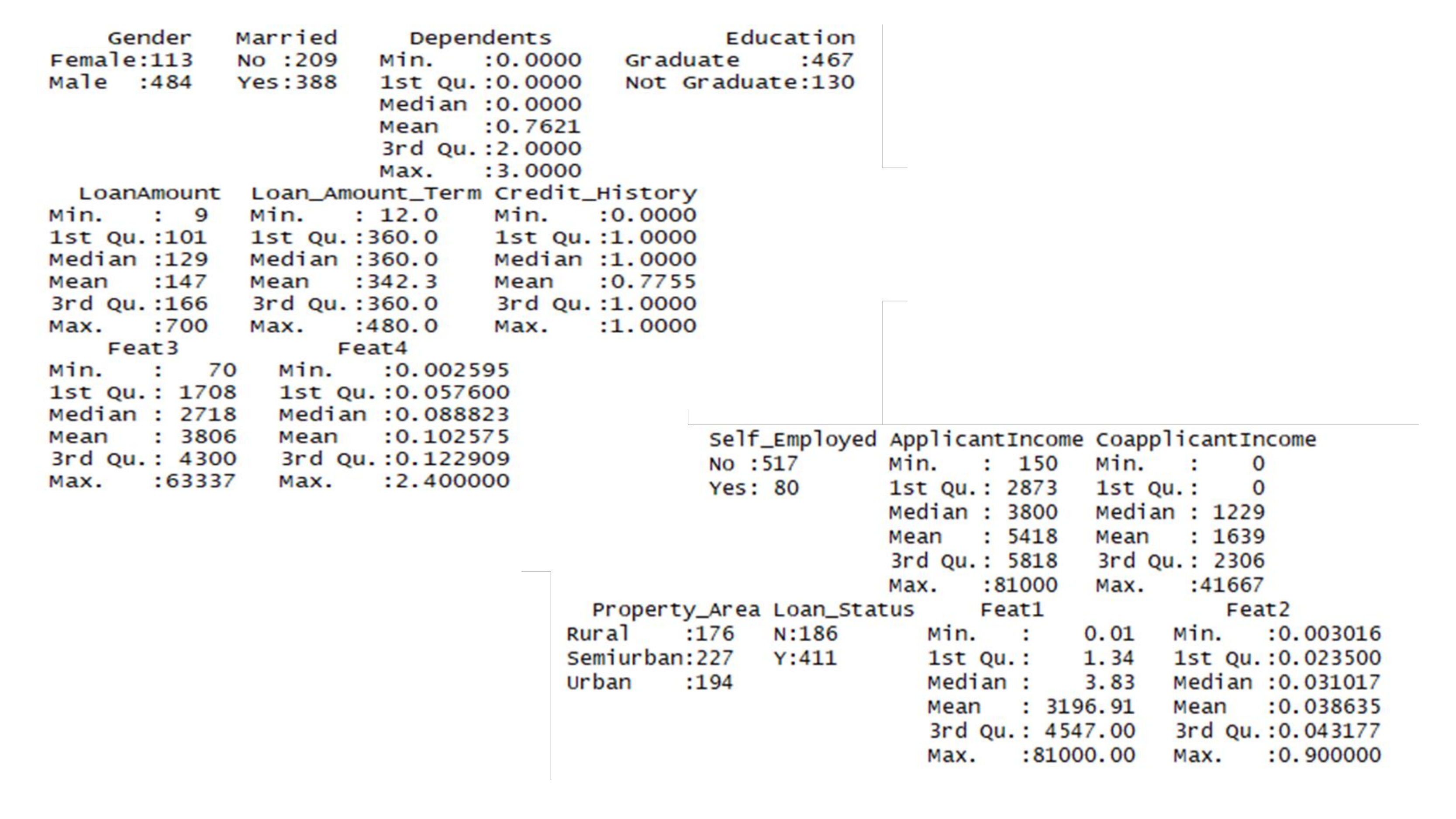}
\caption{This table provides the descriptive statistics of the "Gender Set". This data set contains 597 data points; 113 of these represent women and 484, man. 31\% of the applications contained in the data set have lead to a refusal.}
\label{fig:2}
\end{figure}

\noindent Two data sets used for scoring purposes provided by financial institutions are used in what follows. These data sets contain information about both gender and ethnicity. These data sets are publicly available on either the Kaggle website or UCL's Github. Figures \ref{fig:1} and \ref{fig:2} provide numerous statistical pieces of information regarding the fields of each data set, such as distributions, number of elements per category, and so on. It is important to mention that though one of the data set contains both ethnicity and gender, we opted for two different data sets to ensure the robustness of our analysis by not relying on a single set of information.\\

\subsection{The Ethnicity Set}

\begin{figure}[h!]
\centering
\includegraphics[width=1\textwidth]{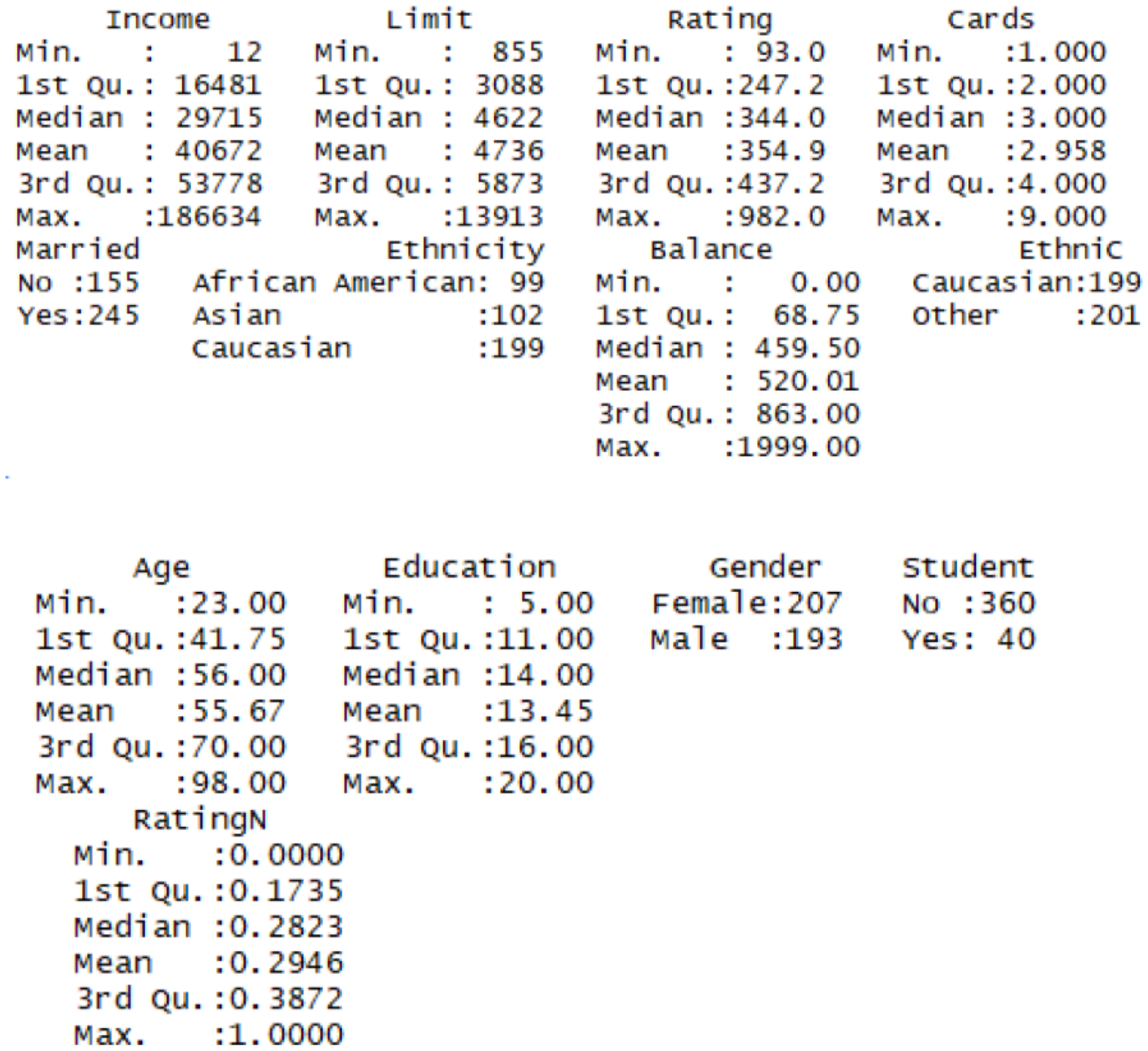}
\caption{This table provides the descriptive statistics of the "Ethnicity Set". The data set contains information about the income of the applicants, current rating, credit limit, the number of credit cards they possess, age, level of education, gender, marital status, ethnicity, and current balance. This data set contains 400 data points. Among these 400 data points, 99 are classified as "African-American", 102 are classified as "Asian" and 199 as "Caucasian". The sample age range from 23 to 98 are are fairly split. Roughly half the sample represents women (207) and the other half men (193).}
\label{fig:1}
\end{figure}

\noindent The first data set, referred to as the \say{Ethnicity Set}\footnote{The data are available at https://www.kaggle.com/suzanaiacob/predicting-credit-card-balance-using-regression}, contains information about the income of the applicants, current rating, credit limit, the number of credit cards they possess, age, level of education, gender, marital status, ethnicity, and current balance. This data set contains 400 data points. Among these 400 data points, 99 are classified as \say{African-American}, 102 are classified as \say{Asian} and 199 as \say{Caucasian}. The sample age range from 23 to 98 are are fairly split. Roughly half the sample represents women (207) and the other half men (193). Figure \ref{fig:1} provides detailed information pertaining to each field included in the data set.\\

\noindent In the considered sample, the average income for African Americans is 44698.37, Asians is 40144.45 and Caucasians 38939.95 dollars. The quartiles representing the income distribution of each ethnic group are represented in Table \ref{table:2}.\\

\begin{table}[h!]
\centering
\begin{tabular}{|c c c c c|} 
 \hline
 0\%  &   25\%  &   50\%  &   75\%   &  100\% \\  
 \hline\hline
 \multicolumn{5}{c}{African American}\\
 \hline
 1409 &  19445 &  33017 &  54860 & 186634  \\ 
 \hline
 \multicolumn{5}{c}{Asian}\\
 \hline
 177 &  15514 &  27732 &  52958 & 180379  \\ 
 \hline
 \multicolumn{5}{c}{Caucasian}\\
 \hline
 12 &  16293 & 30002 &  53943 & 182728   \\  
 \hline
\end{tabular}
\caption{This table presents the quartiles of African American, Asian and Caucasian distributions of income.}
\label{table:2}
\end{table}

\begin{figure}[h!]
\centering
\includegraphics[width=1\textwidth]{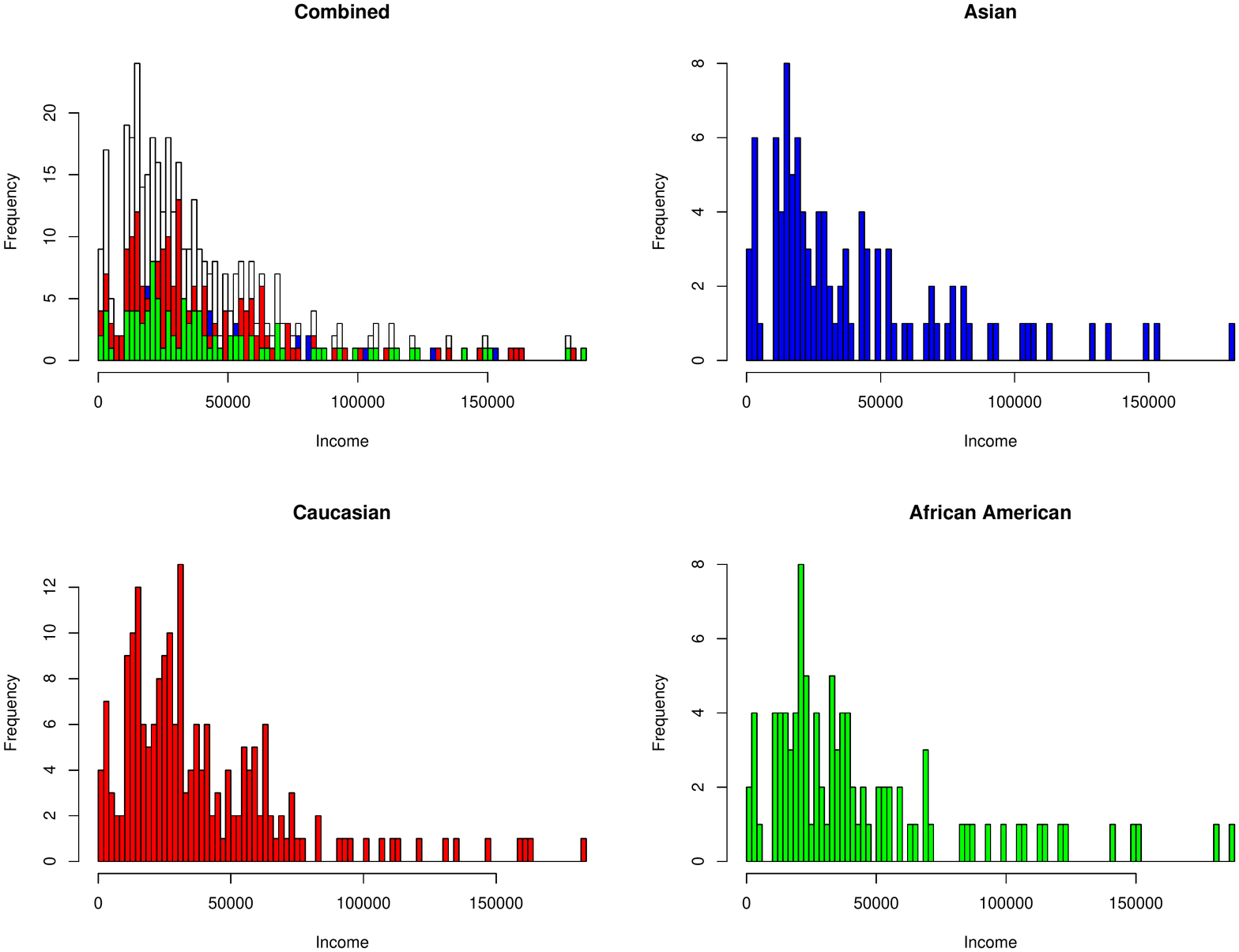}
\caption{This figure presents four histograms. On the top left end corner, the various group are represented simultaneously. The distribution of income of the whole data set is depicted along the distribution of income of each ethnic group. The three other histograms depict the distribution of income for each ethnic group, i.e. Caucasian, African American and Asian.}
\label{fig:3}
\end{figure}

\begin{figure}[h!]
\centering
\includegraphics[width=1\textwidth]{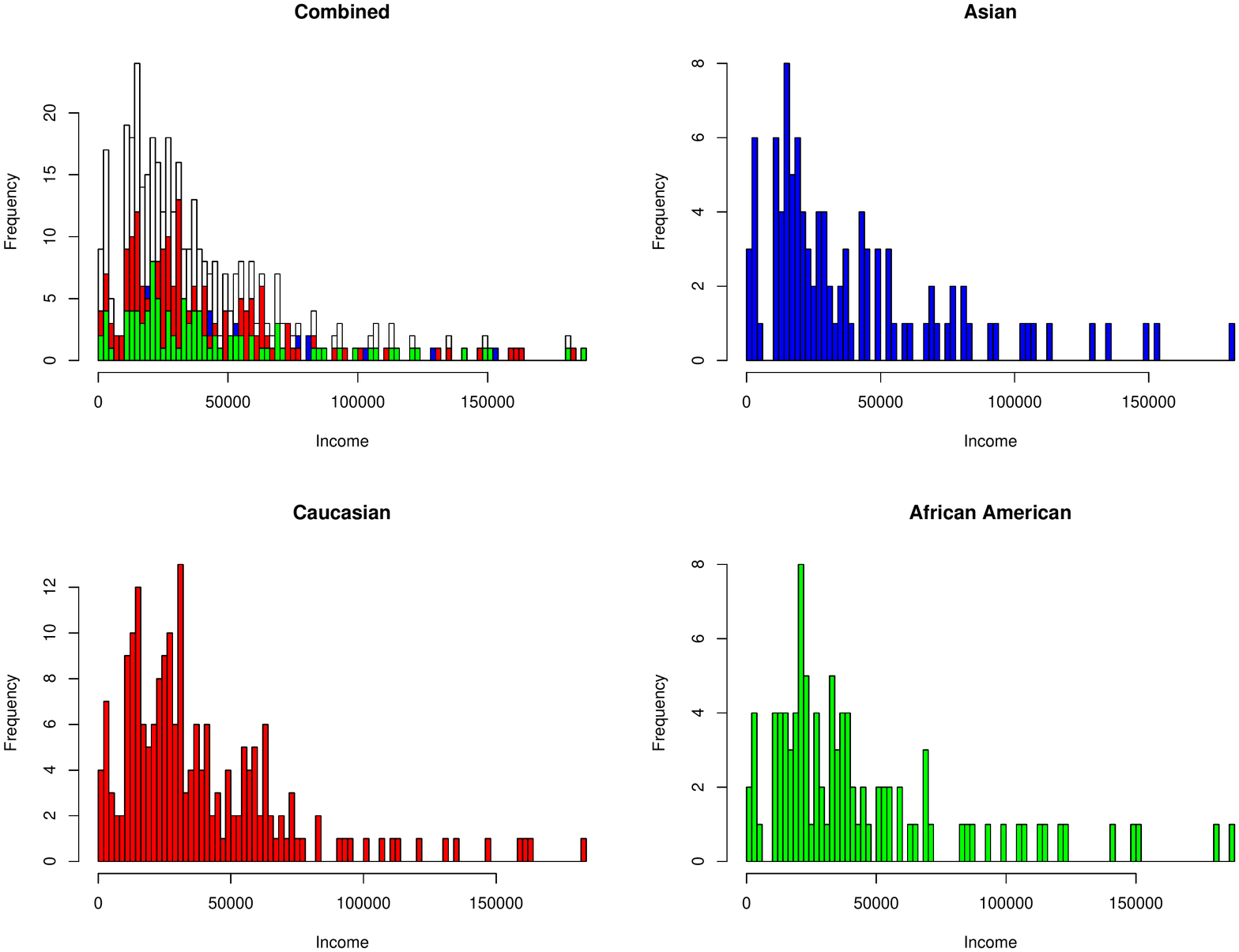}
\caption{This figure presents four histograms. These histograms have been obtained using altered data. On the top left end corner, four histograms are represented simultaneously, showing the distribution of income in the data set along with the distribution of income of each ethnic group. The three other histograms depict the distribution of income for each ethnic group, i.e. Caucasian, African American and Asian.}
\label{fig:55}
\end{figure}

\noindent We see in the ethnicity data that the three groups are having fairly similar distributions of income. As such, it is not representative of what has been reported in the various reports aforementioned. Therefore, after working on the data set as obtained, we will also analyse the impact of modifying the income of these groups changing the income by a certain coefficient in order to better reflect what has been reported by NGOs, as such we will create an alternate data set were African Americans are earning 25\% less than Caucasians and Asians 10\% less than Caucasians bringing the average down to 33523.78 for African Americans and 36130.01 for Asians. In Table \ref{table:3}, the quartiles of the modified data are provided. Figure \ref{fig:3} and Figure \ref{fig:55} depict the histogram respectively related to the original \say{Ethnicity Set} and the modified one.   

\begin{table}[h!]
\centering
\begin{tabular}{|c c c c c|} 
 \hline
 0\%  &   25\%  &   50\%  &   75\%   &  100\% \\ 
 \hline\hline
 \multicolumn{5}{c}{African American}\\
 \hline
 1057 &  14583 &  24763 &  41145 & 139976  \\
 \hline
 \multicolumn{5}{c}{Asian}\\
 \hline
 159 &  13963 &  24959 &  47662 & 162341  \\ 
 \hline
 \multicolumn{5}{c}{Caucasian}\\
 \hline
 12 & 16293 &  30002 &  53943 & 182728    \\
 \hline
\end{tabular}
\caption{This table presents the quartiles of African American, Asian and Caucasian distributions of income, after alteration of the data set.}
\label{table:3}
\end{table}

\subsection{The Gender Set}

\noindent The second data set, referred to as the \say{Gender Set}\footnote{The data are available at https://www.kaggle.com/ajaymanwani/loan-approval-prediction https://github.com/shrikant-temburwar/Loan-Prediction-Dataset}, contains information about the gender of the applicants, marital status, whether they have dependents, level of education, whether they are self employed or not, income, income of the co-applicant, the amount of the loan requested, the term of the requested loan, credit history, the location of their current property and the status of their loan. This data set contains 597 data points; 113 of these represent women and 484, man. 31\% of the applications contained in the data set have led to a refusal.\\

\begin{figure}[h!]
\centering
\includegraphics[width=1\textwidth]{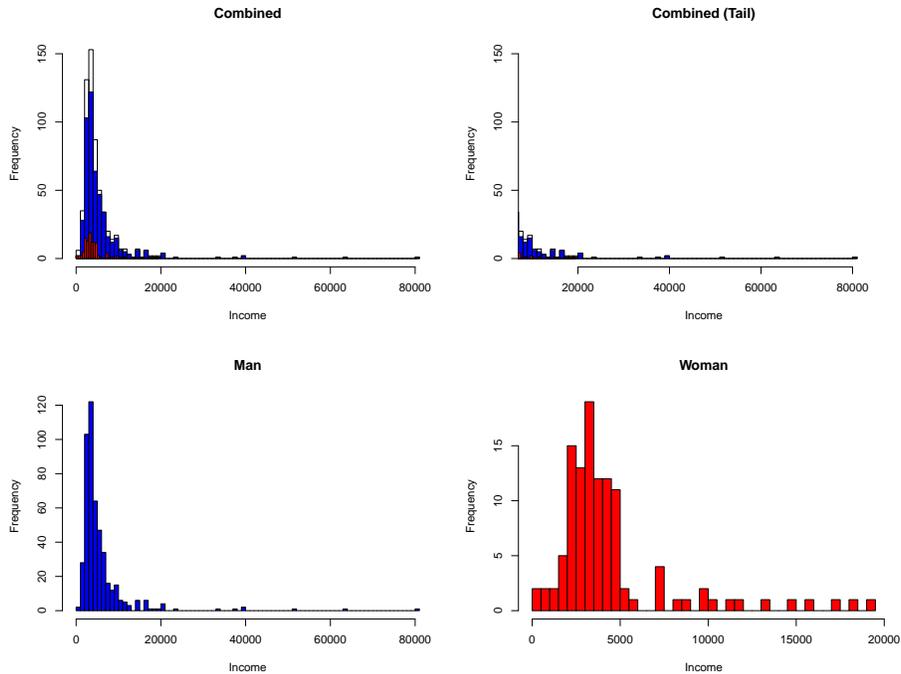}
\caption{This figure presents four histograms. On the top left end corner, the various groups are represented simultaneously. The distribution of income of the whole data set is depicted along the distribution of income of each gender. Two other histograms depict the distribution of income of each gender group (bottom left and bottom right). The histogram located in the top right-hand corner represents the tail of the income distribution, showing that over a certain threshold, women are not represented anymore.}
\label{fig:4}
\end{figure}

\noindent In Figure \ref{fig:4}, the income by gender has been represented; as can be observed, the sample is consistent with what has been reported internationally; there is a clear gap in terms of remuneration, and women are clearly earning less than men on average. Unfortunately, since the type of employment is not shown, it is not possible to investigate the matter further, but there is no reason why this should affect our reasoning, as any inequality would be reflected accordingly. \noindent Indeed, the average monthly income of women in the data set considered, we observed an average of 4530.468 dollars, while for men, this average went up to 5769.968 dollars, i.e. a difference of 27.36\%. The quantiles representing the income distribution are provided in Table \ref{table:1}.\\

\begin{table}[h!]
\centering
\begin{tabular}{|c c c c c|} 
 \hline
 0\%  &   25\%  &   50\%  &   75\%   &  100\% \\  
 \hline
 \hline
 \multicolumn{5}{c}{Women}\\
 \hline
 210 &  2870 &  3655 & 4727 & 19484 \\
 \hline
 \multicolumn{5}{c}{Men}\\
 \hline
 150 & 2980 & 3859 & 5827 & 81000 \\
 \hline
\end{tabular}
\caption{This table presents the quartiles of both women and men distributions of income.}
\label{table:1}
\end{table}

\section{Methodology}

\noindent In this paper, we assume that for a credit scoring data set to be unbiased, the information provided should not contain any direct or indirect information susceptible to give away the gender or the ethnic group of customers. Therefore, our main objective is to try to figure out or predict either the gender or the ethnicity of customers based on data used for credit scoring purposes. Though this paper would gain from being tested on larger or different data sets, the results obtained implementing the following approaches are easily extendable. Furthermore, it is worth mentioning that though in most countries the ethnicity of customers is not given, if the data contains information characteristic of a certain group (for instance the level of remuneration), then not having an explicit field does not solve the problem. However, the fact that a field explicitly either state the gender or the ethnicity of the customer permits testing our hypothesis. In what follows, we will proceed in three steps:

\begin{enumerate}

\item The first step is to test whether the data are actually usable for credit scoring purposes. In other words, we are going to test if it is possible to perform a regression to predict the scores using the \say{Ethnicity Set} and if it is possible to perform a classification to predict whether their application will be approved or not using the \say{Gender Set}.

\item In a second step, we will try to predict either the gender or the ethnicity of the customers contained in the database.

\item In a third step, we will try to improve the prediction.
\end{enumerate}

\noindent When the variable to be predicted is continuous, we will perform a regression. When the response variable is discreet, we will perform a classification. A similar algorithm can be used in both situations. Following \cite {guegan2018regulatory} and \cite{addo2018credit}, we initially used a random forest growing 750 trees. Random forests operate by constructing a multitude of decision trees at training time and producing the class as the output according to the mode of the classes or the mean prediction of each individual trees respectively. Random forests correct for decision trees overfitting tendency \cite{breiman2001random}.\\ 

\noindent To evaluate the quality of the regression we will use the mean squared error (mse) and for the classifications the F1-Score which is equal to $2 \times \frac{precision \times recall}{precision + recall}$.

\subsection{Ethnicity Set}

\noindent For the \say{Ethnicity Set}, in a first step, we will assess the suitability of the data set for scoring purposes. Therefore, the sample is split in two subsamples; 75\% of the initial set is used for training purposes and 25\% for testing purposes. As the response variable is continuous, to assess the suitability of the data set we will perform a regression. The mse obtained is equal to 0.008210515, supporting the conclusion that the data set is adequate for scoring purposes.\\

\begin{figure}[h!]
\centering
\includegraphics[width=1\textwidth]{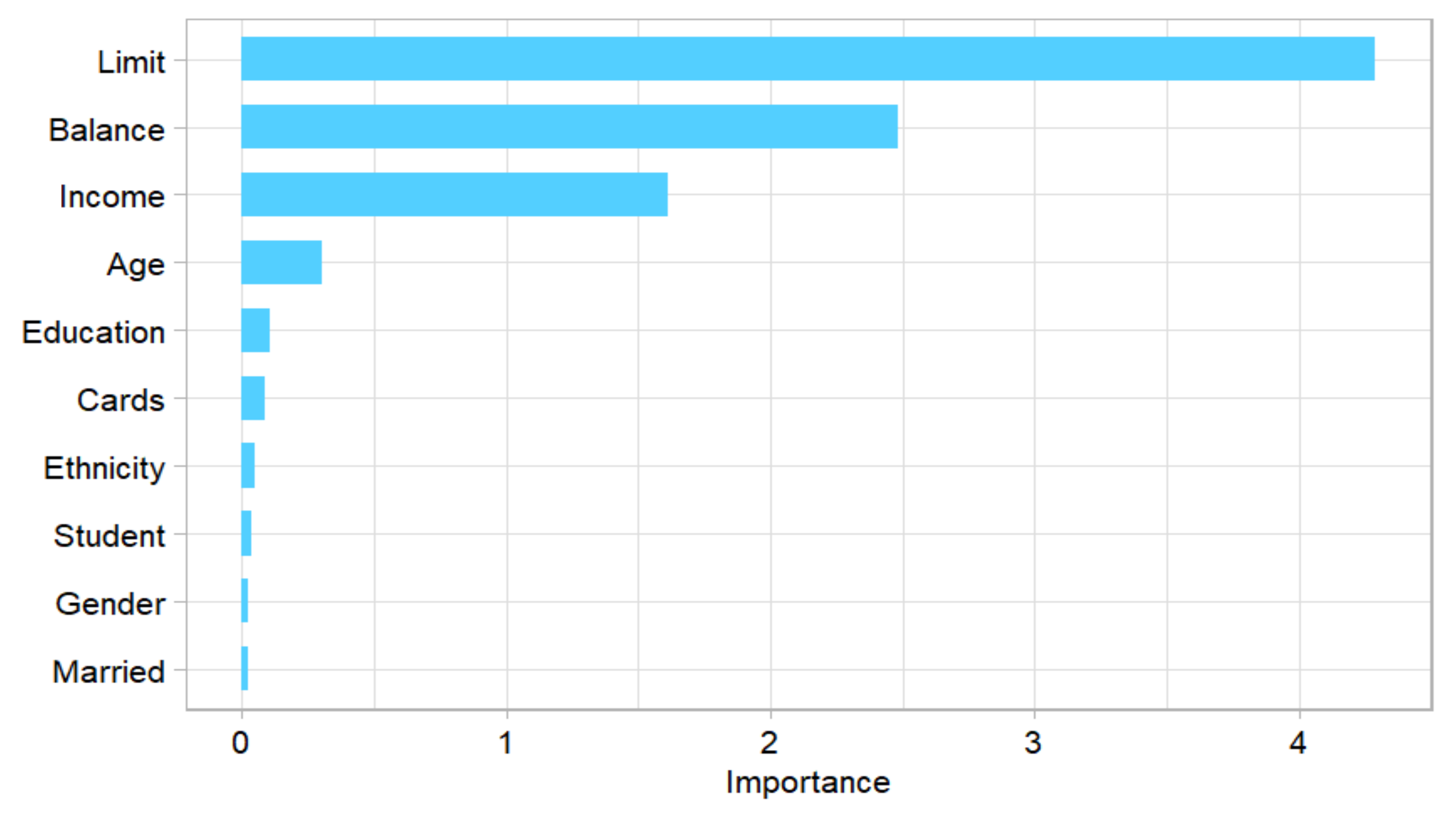}
\caption{This figure presents the graph of "variable importance" for the "Ethnicity Set". It is interesting to note that the graph confirms the fact that the three most important variables are all related to the financial wealth of customers.}
\label{fig:6}
\end{figure}

\noindent In a second step, we will now be using an identical data set to predict the ethnic group of the customers, facing now a classification problem, we obtained a F1-score equal to 0.6507937. This result demonstrates that the data are already containing a lot of information regarding the ethnic affiliation of the bank's customers. Figure \ref{fig:6} also provides the weight of each variable in the predictions, and it appears that the factors related to the financial wealth of the applicants are predominant, i.e., the current credit limit, the money available on their bank account and their income. Thus, it is not surprising that people earning less money face a lower access to credit.\\

\noindent In a third step, to further test our hypothesis, we will try to predict the ethnic group of each customer contained in the data set after modifying the data to better reflect the reality. After modifying the revenues of the different ethnic groups as well as the related elements such as their ratings, when we tried to reclassify, the results were spectacular, the quality of the classification as given by the F1-score was 0.7, and went up 0.9863014 when we implemented an oversampling strategy to rebalance the data set, i.e. creating synthetic data points in such a way that the three ethnic groups are represented by a population of similar sizes (see Table \ref{table:Ethnicity}).\\

\noindent As a conclusion, the information transmitted to financial institutions when applying for a loan contain sufficient information to figure out the ethnic group of the customers, and the pertaining biases mechanically transmitted into their evaluation.\\

\begin{table}[h!]
\centering
\begin{tabular}{|c | c |} 
 \hline
 Data as provided & 0.6507937\\ 
 \hline
 Data Modified  & 0.7 \\
 \hline
 Data Modified Smote & 0.9863014\\
 \hline
\end{tabular}
\caption{This table presents the F1-score obtained for the random forest classification performed using the ''Ethnicity Set'' for ethnicity prediction purposes.}
\label{table:Ethnicity}
\end{table}

\subsection{Gender Set}

\begin{figure}[h!]
\centering
\includegraphics[width=1\textwidth]{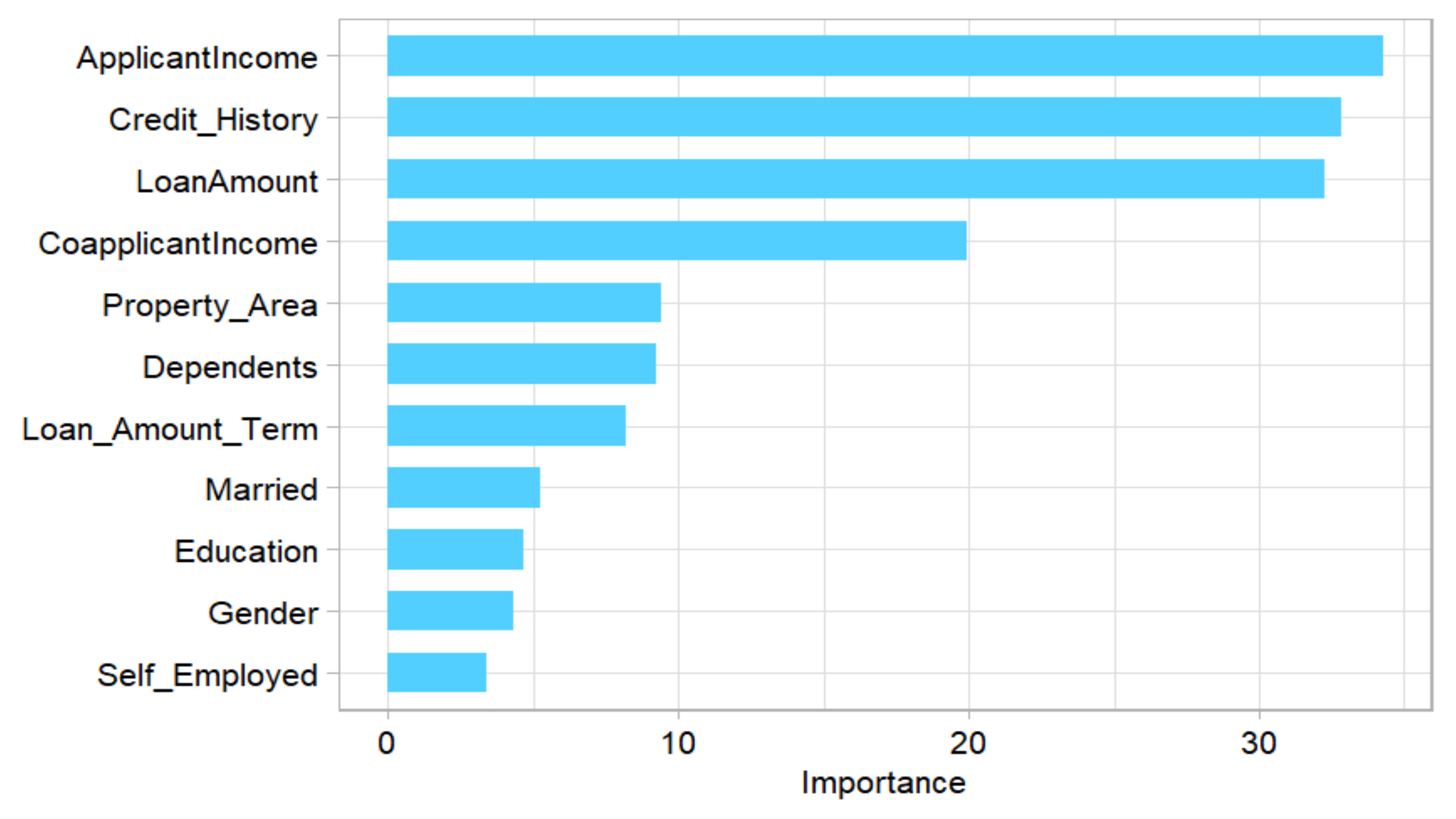}
\caption{This figure presents the graph of ''variable importance'' for the ''Gender Set''. As for the "Ethnicity Set" , the most important variables are related to customers financial wealth.}
\label{fig:ImpGender}
\end{figure}

\noindent As for the \say{Ethnicity Set}, the \say{Gender Set} was split: 75\% of the initial set was used for training purposes and 25\% for testing purposes. Once again, in the first step, we checked if the data set was adequate for credit scoring purposes. The results regarding the loan approval predictions are provided in Table \ref{table:Ethnicity}. The initial F1-score obtained is equal to 0.5052632 which is not sufficient to validate the hypothesis. We assumed that feature engineering might improve the algorithm performance, but once again the F1-score obtained was equal to 0.5, which is not sufficient to validate this subsequent hypothesis. After further investigation, we noticed that the data set was unbalanced, i.e., there was a lot more approvals than refusals (however, not unbalanced enough to provide unreliable results) in the data set. To overcome that issue, we implemented a SMOTE strategy allowing rebalancing the data set. The SMOTE approach was implemented to increase the size of the information set related to unapproved loans. Following this procedure, the F1-score increased to 0.8295189. Adding up feature engineering, the F1-score went up to 0.843418 (see Table \ref{table:GenderLoan}). Therefore, the data set can be used for scoring purposes. Figure \ref{fig:ImpGender} presents the \say{variable importance} graph showing that on this data set, the applicant income is one of the main factor driving the results.\\

 \begin{table}[h!]
\centering
\begin{tabular}{|c | c |} 
 \hline
 & \textbf{Random Forest} \\
 \hline
 Data as provided &  0.5052632 \\
 \hline
 Features Engineered & 0.5 \\
 \hline
 Smote & 0.8295189 \\
 \hline
 Smote Features Engineered & 0.843418\\
 \hline
\end{tabular}
\caption{This table presents the F1-Score obtained for the random forest classification performed using the "Gender Set" for loan approval prediction purposes. Once the resampling strategy (SMOTE) has been applied, the performance of the algorithm is sufficient to precisely predict customers' loan approvals.}
\label{table:GenderLoan}
\end{table}

\noindent Considering the prediction of customers' gender, the results are following the same patterns. The F1-score obtained on the raw data is equal to 0.3333333. To  improve the quality of the adjustments, the following features have been engineered:
\begin{enumerate}
\item Applicant Income / (Coapplicant Income + 1)
\item Loan Amount / Applicant Income
\item Applicant Income / (Dependents + 1)
\item Loan Amount Term / Applicant Income
\end{enumerate}
\noindent Using feature engineering, the result of the F1-score is equal to 0.3666667. However, with the SMOTE approach, the result increased to 0.8583765, and to 0.8773748 (see \ref{table:Gender}), once the features engineered had been added. Therefore, the same data set can be used for gender prediction purposes.\\

 \begin{table}[h!]
\centering
\begin{tabular}{|c | c |} 
 \hline
 & \textbf{Random Forest} \\
 \hline
 Data as provided  & 0.3333333 \\ 
 \hline
 Features Engineered & 0.3666667 \\
 \hline
 Smote & 0.8583765 \\
 \hline
 Smote Features Engineered & 0.8773748\\ 
 \hline
\end{tabular}
\caption{This table presents the F1-score obtained for the random forest classification performed using the ''Gender Set'' for gender prediction purposes. Once the resampling strategy (SMOTE) has been applied, the performance of the algorithm is sufficient to precisely predict customers' gender.}
\label{table:Gender}
\end{table}

\section{Conclusion}

\noindent In this paper, our objective was to assess if social biases were captured into credit scoring, and the assumption we made was that if social biases were not included, then factors characterising credit scores would be sufficiently different from those that can characterize either men, women or any ethnic groups. If the data used to score customers can be used to predict any sensitive information, and if the data are socially biased then the credit score will also be biased.\\

\noindent The most interesting part of the analysis is the fact that results obtained to score customers can be used to predict if the gender or the ethnicity of the customers and thus all social biases translated in the data are mechanically included in the scores, and therefore, discrimination is mechanically translated into loan supply, and kept in the data sets for training purposes ensuring that such discrimination continues and is potentially reinforced. Through that mechanism, social biases become societal biases, as driven by the norm.\\

\noindent What is quite interesting is that it could be possible to unbias the datasets, however, if we consider that a customer with a lower revenue is riskier for a bank than a customer with a higher revenue, then correcting the biases by ensuring that social biases are not captured in the data could lead financial institutions to take higher risks. Thus, one may wonder if the solution would not come from the regulator itself. Another aspect appeared in this analysis, if the data set is homogeneous, it becomes complicated to predict either the gender or the ethnic groups though it would still be possible to score the customer. Unfortunately, this might lead to fully unbalanced subsamples in which we would have non approved loans on one side and approved loans on the other. Unbiasing either the data set or the algorithm will be the topic of our next paper, though we will have to address the issue carefully considering that unbiasing a data set is likely to engender an opposite bias.

\newpage

\bibliographystyle{apalike}
\bibliography{Bias}

\begin{thebibliography}{}

\bibitem[Addo et~al., 2018]{addo2018credit}
Addo, P.~M., Guegan, D., and Hassani, B. (2018).
\newblock Credit risk analysis using machine and deep learning models.
\newblock {\em Risks}, 6(2):38.

\bibitem[Breiman, 2001]{breiman2001random}
Breiman, L. (2001).
\newblock Random forests.
\newblock {\em Machine learning}, 45(1):5--32.

\bibitem[Guegan and Hassani, 2018]{guegan2018regulatory}
Guegan, D. and Hassani, B. (2018).
\newblock Regulatory learning: how to supervise machine learning models? an
  application to credit scoring.
\newblock {\em The Journal of Finance and Data Science}, 4(3):157--171.

\bibitem[Hall et~al., 2016]{hall2016black}
Hall, A.~V., Hall, E.~V., and Perry, J.~L. (2016).
\newblock Black and blue: Exploring racial bias and law enforcement in the
  killings of unarmed black male civilians.
\newblock {\em American Psychologist}, 71(3):175.

\bibitem[Hegewisch and Hartmann, 2018]{hegewisch2018gender}
Hegewisch, A. and Hartmann, H. (2018).
\newblock The gender wage gap: 2018; earnings differences by race and
  ethnicity.
\newblock {\em Institute for Women’s Policy Research}, 7.

\bibitem[Holmes, 2019]{CreditCards2019}
Holmes, T.~E. (2019).
\newblock Credit card race, age, gender statistics.
\newblock {\em creditcards.com}.

\bibitem[Levinson and Smith, 2012]{levinson2012implicit}
Levinson, J.~D. and Smith, R.~J. (2012).
\newblock {\em Implicit racial bias across the law}.
\newblock Cambridge University Press.

\bibitem[ONS, 2019]{ons2019}
ONS (2019).
\newblock Ethnicity pay gaps in great britain: 2018.
\newblock {\em Office of National Statistics, United Kingdom}.

\bibitem[Witzany, 2017]{witzany2017credit}
Witzany, J. (2017).
\newblock Credit risk management.
\newblock In {\em Credit Risk Management}, pages 5--18. Springer.

\end{thebibliography}

\end{document}